\documentclass[11pt]{article}

\usepackage[utf8]{inputenc}
\usepackage[T1]{fontenc}
\usepackage{amsmath,amssymb,amsfonts}
\usepackage{graphicx}
\usepackage{booktabs}
\usepackage{multirow}
\usepackage{hyperref}
\usepackage{xcolor}
\usepackage{algorithm}
\usepackage{algorithmic}
\usepackage{enumitem}
\usepackage{subcaption}
\usepackage{tabularx}
\usepackage{float}
\usepackage{tikz}
\usetikzlibrary{arrows.meta,decorations.pathreplacing}
\usepackage[margin=1in]{geometry}
\usepackage{natbib}
\usepackage{listings}
\usepackage{tcolorbox}

\definecolor{teal}{RGB}{0,128,128}
\definecolor{codeblue}{RGB}{41,128,185}
\definecolor{codegray}{RGB}{128,128,128}
\definecolor{codegreen}{RGB}{39,174,96}
\definecolor{codepurple}{RGB}{142,68,173}
\definecolor{backcolor}{RGB}{248,248,248}

\lstdefinestyle{mystyle}{
    backgroundcolor=\color{backcolor},
    commentstyle=\color{codegray},
    keywordstyle=\color{codeblue},
    stringstyle=\color{codegreen},
    basicstyle=\ttfamily\footnotesize,
    breaklines=true,
    frame=single,
    rulecolor=\color{codegray!30},
    numbers=none,
}
\newcommand{\byterover}{\textsc{ByteRover}}
\newcommand{\contexttree}{\textit{Context Tree}}

\title{\textbf{ByteRover: Agent-Native Memory Through \\ LLM-Curated Hierarchical Context}}

\author{
  Andy Nguyen\textsuperscript{1},
  Danh Doan\textsuperscript{1},
  Hoang Pham\textsuperscript{1},
  Bao Ha\textsuperscript{1},
  Dat Pham\textsuperscript{1},
  Linh Nguyen\textsuperscript{1},\\
  Hieu Nguyen\textsuperscript{1},
  Thien Nguyen\textsuperscript{1},
  Cuong Do\textsuperscript{1},
  Phat Nguyen\textsuperscript{1},
  Toan Nguyen\textsuperscript{1} \\[6pt]
  \textsuperscript{1}ByteRover \\
  \url{https://www.byterover.dev}
}

\date{}

\begin{document}
\maketitle

\begin{abstract}
Memory-Augmented Generation (MAG) extends large language models with external memory to support long-context reasoning,
but existing approaches universally treat memory as an \emph{external service} that agents call into---delegating
storage to separate pipelines of chunking, embedding, and graph extraction.
This architectural separation means the system that stores knowledge does not understand it,
leading to semantic drift between what the agent intended to remember and what the pipeline actually captured,
loss of coordination context across agents, and fragile recovery after failures.
In this paper, we propose \byterover{}, an \emph{agent-native} memory architecture that inverts
the memory pipeline: the same LLM that reasons about a task also curates, structures, and retrieves knowledge.
\byterover{} represents knowledge in a hierarchical \contexttree{}---a file-based knowledge graph organized as
\textit{Domain~$>$~Topic~$>$~Subtopic~$>$~Entry}---where each entry carries explicit relations,
provenance, and an Adaptive Knowledge Lifecycle (AKL) with importance scoring, maturity tiers, and recency decay.
Retrieval uses a 5-tier progressive strategy that resolves most queries at sub-100\,ms latency
without LLM calls, escalating to agentic reasoning only for novel questions.
Experiments on LoCoMo and LongMemEval demonstrate that \byterover{} achieves
state-of-the-art accuracy on LoCoMo and competitive results on LongMemEval
while requiring \emph{zero external infrastructure}---no vector database,
no graph database, no embedding service---with all knowledge stored as human-readable markdown files on the local filesystem.
\end{abstract}

\section{Introduction}
\label{sec:intro}

Large Language Models (LLMs) have demonstrated remarkable capabilities across a wide range of tasks~\citep{brown2020language,achiam2023gpt4,wei2022chain},
yet they remain fundamentally limited in their ability to maintain and reason over long-term context.
These models process information within a finite attention window,
and their internal representations do not persist across interactions,
causing earlier details to be forgotten once they fall outside the active context~\citep{brown2020language,beltagy2020longformer}.
Even within a single long sequence, attention effectiveness degrades with distance
due to \emph{attention dilution}, positional encoding limitations, and token interference,
leading to the well-known ``lost-in-the-middle'' phenomenon~\citep{liu2024lost,press2021train}.

To address these limitations, Memory-Augmented Generation (MAG) systems have emerged
as a promising direction for enabling LLMs to operate beyond the boundaries of their fixed context windows~\citep{xu2025amem,nan2025nemori,jiang2026magma,chhikara2025mem0}.
MAG equips an agent with an external memory module that continuously records interaction histories
and allows the agent to retrieve and reintegrate past experiences when generating new responses.
The paradigm has rapidly evolved from lightweight semantic stores to entity-centric,
episodic, and hierarchical designs~\citep{jiang2026anatomy}.

Despite this architectural diversity, current MAG systems share a common structural pattern:
\textbf{memory is an external service that agents call into}.
The agent serializes data, sends it to a separate pipeline (for chunking, embedding, entity extraction, or graph construction),
receives an acknowledgment, and later queries the service to retrieve results.
The pipeline that stores knowledge does not understand it---chunking is mechanical,
embeddings encode surface similarity rather than semantic relationships,
and the agent has no visibility into how its memories were organized or why certain relationships were created.

This external-service paradigm creates three failure modes
that become critical for autonomous agents operating over long time horizons:

\begin{enumerate}[leftmargin=*]
  \item \textbf{Semantic drift.}
    The agent's understanding of what it stored diverges from what the memory service actually captured.
    The agent intends to store a nuanced insight; the pipeline chunks and embeds it differently,
    and the next retrieval returns a tangentially related fragment.

  \item \textbf{Lost coordination context.}
    When multiple agents share an external memory service, they share data but not understanding.
    Agent~A stores a finding with reasoning and rationale.
    Agent~B retrieves the data but lacks the \emph{why}---what reasoning led to the conclusion,
    what actions were expected to follow.
    The provenance is lost in the embedding.

  \item \textbf{Recovery fragility.}
    When an autonomous agent crashes mid-task, it must reconstruct state by querying the memory service,
    interpreting results, and inferring where it left off.
    With stateful, file-based memory, the state is \emph{in the files}---per-operation status,
    timestamps, and the knowledge structure itself tells the agent exactly what was completed.
\end{enumerate}

To address these limitations, we propose \byterover{}, an \emph{agent-native} memory architecture
that inverts the relationship between agent and memory.
Instead of calling an external memory service, the same LLM that reasons about a task
also curates knowledge into a hierarchical \contexttree{}---a file-based knowledge graph
where each entry carries explicit relations, provenance, and lifecycle metadata.
Memory operations (ADD, UPDATE, UPSERT, MERGE, DELETE) are tools in the agent's toolkit,
not API calls to an external service, enabling a stateful feedback loop
where the agent sees per-operation results and adapts in real time.

Our contributions are summarized as follows:

\begin{enumerate}[leftmargin=*]
  \item We propose \byterover{}, an agent-native memory architecture where the LLM itself curates knowledge
    through structured operations with explicit provenance and rationale,
    eliminating the separation between the ``understanding'' agent and the ``storing'' pipeline.

  \item We introduce the \contexttree{}, a hierarchical file-based knowledge graph
    with an Adaptive Knowledge Lifecycle (AKL)---importance scoring, maturity tiers (draft~$\to$~validated~$\to$~core),
    and recency decay---that enables knowledge to naturally evolve over time.

  \item We design a 5-tier progressive retrieval strategy that resolves most queries
    at sub-100\,ms latency without LLM calls, combined with out-of-domain detection
    that explicitly signals when queries fall outside stored knowledge.

  \item We demonstrate that \byterover{} achieves state-of-the-art results on LoCoMo and competitive
    results on LongMemEval benchmarks while requiring zero external infrastructure---no vector database, no graph database,
    no embedding service---with all knowledge stored as human-readable markdown files.
\end{enumerate}

\section{Background}
\label{sec:background}

\subsection{Memory-Augmented Generation}

Agentic memory extends retrieval-based generation by introducing persistent, writable memory
that evolves across interactions~\citep{jiang2026anatomy}.
Formally, at step~$t$, the agent conditions on observations~$o_t$ and an external memory state~$\mathcal{M}_t$:
\begin{equation}
  y_t \sim f_\theta\!\Big(\phi(o_t, s_t) \;\oplus\; \psi(\mathcal{M}_t; q_t)\Big),
  \label{eq:mag}
\end{equation}
where $y_t$ denotes the output, $s_t$ is additional agent state,
$\psi(\mathcal{M}_t; q_t)$ retrieves memory given query~$q_t$,
and $\oplus$ represents integration (e.g., prompt concatenation).
Crucially, memory affects behavior through the explicit retrieval term~$\psi$
rather than updates to~$\theta$.

Two coupled processes are operated: \textbf{inference-time recall}
(reading memory to condition decisions) and \textbf{memory update}
(writing, consolidating, and forgetting to maintain a useful long-term store).
The memory module evolves via a feedback loop:
\begin{align}
  o_t &= \text{LLM}(q_t, \text{Retrieve}(q_t, \mathcal{M}_t)), \\
  \mathcal{M}_{t+1} &= \text{Update}(\mathcal{M}_t, q_t, o_t).
\end{align}

\subsection{Taxonomy of Existing Approaches}

Following the taxonomy introduced by \citet{jiang2026anatomy},
existing MAG systems can be organized into four structural categories:

\begin{itemize}[leftmargin=*]
  \item \textbf{Lightweight Semantic Memory.}
    Independent textual units embedded in a vector space and retrieved via top-$k$ similarity.
    No explicit structural relations~\citep{liu2026simplemem}.

  \item \textbf{Entity-Centric and Personalized Memory.}
    Organizes information around explicit entities using structured records
    or attribute-value pairs~\citep{xu2025amem,modarressi2023retllm,chhikara2025mem0}.

  \item \textbf{Episodic and Reflective Memory.}
    Adds temporal abstraction by organizing interactions into episodes
    or higher-level summaries~\citep{nan2025nemori}.

  \item \textbf{Structured and Hierarchical Memory.}
    Imposes explicit organization over stored information via
    graphs~\citep{jiang2026magma,rasmussen2025zep},
    hierarchical tiers~\citep{kang2025memoryos,packer2023memgpt},
    or policy-optimized management.
\end{itemize}

\subsection{The External Service Paradigm}
\label{sec:external-service}

Despite their architectural diversity, \emph{all} systems in the taxonomy above
share a common interaction pattern: the agent communicates with memory through an API boundary.
The agent serializes data, the memory service processes it through its own pipeline
(chunking, embedding, entity extraction, graph construction),
and the agent later queries the service to retrieve results.

This pattern has a fundamental consequence:
\textbf{the system that stores knowledge does not understand it.}
The embedding model that creates vector representations operates independently of the agent's reasoning.
The entity extraction pipeline has its own notion of what matters.
The agent has no visibility into how its memories were organized, why certain relationships were created,
or whether the stored representation faithfully captures its intent.

\byterover{} departs from this paradigm by making the agent itself the curator---the
same LLM that reasons about a task decides what to store, where to place it,
what it relates to, and why it matters.

\section{The \contexttree{}}
\label{sec:design}

This section introduces the \byterover{} architecture and its core data structure, the \contexttree{}.

\subsection{Architectural Overview}
\label{sec:arch-overview}

\begin{figure}[t]
  \centering
  \begin{tikzpicture}[
    font=\small,
    box/.style={draw, rounded corners=2pt, minimum height=0.55cm,
                text centered, inner sep=3pt, font=\scriptsize},
    layer/.style={draw, rounded corners=3pt, fill=#1,
                  inner sep=6pt},
    arr/.style={-{Stealth[length=4pt]}, semithick},
    darr/.style={{Stealth[length=4pt]}-{Stealth[length=4pt]}, semithick},
    lbl/.style={font=\scriptsize\sffamily, text=black!50},
  ]


    \node[box, fill=gray!6, minimum width=1.8cm] (cli-tui) at (-2.4,7.2) {TUI};
    \node[box, fill=gray!6, minimum width=1.8cm] (cli-cli) at (0,7.2) {CLI};
    \node[box, fill=gray!6, minimum width=1.8cm] (cli-mcp) at (2.4,7.2) {MCP};

    \node[layer=gray!8, minimum width=8.0cm, minimum height=1.0cm]
      (daemon) at (0,6.0) {};
    \node[font=\scriptsize\bfseries, black!40] at (0,6.35) {Daemon};
    \node[box, fill=white, minimum width=2.6cm] (tq) at (-2.0,6.0)
      {Task Queue};
    \node[box, fill=white, minimum width=2.6cm] (pool) at (2.0,6.0)
      {Agent Pool};

    \draw[arr, gray!50] (cli-tui.south) -- (cli-tui.south |- daemon.north);
    \draw[arr, gray!50] (cli-cli.south) --
      node[right, font=\tiny\sffamily, text=black!50, xshift=1pt] {Socket.IO}
      (cli-cli.south |- daemon.north);
    \draw[arr, gray!50] (cli-mcp.south) -- (cli-mcp.south |- daemon.north);

    \draw[thick, rounded corners=6pt, blue!25]
      (-4.3,5.15) rectangle (4.3,-0.85);
    \node[font=\scriptsize\bfseries, blue!40, anchor=west] at (-4.1,4.9) {Agent Process};
    \node[font=\tiny, blue!25, anchor=east] at (4.1,4.9) {\itshape one per project};

    \draw[arr, gray!50] (pool.south) -- (pool.south |- 0,5.15);

    \node[layer=blue!6, minimum width=8.0cm, minimum height=1.4cm]
      (al) at (0,3.9) {};
    \node[font=\scriptsize\bfseries, blue!50] at (0,4.4) {Agent Layer};
    \node[box, fill=white, minimum width=2.8cm] (llm) at (-2.0,3.7)
      {LLM Loop};
    \node[box, fill=white, minimum width=3.4cm] (mtools) at (2.0,3.7)
      {curate / search\_knowledge};

    \node[layer=green!6, minimum width=8.0cm, minimum height=1.4cm]
      (el) at (0,2.1) {};
    \node[font=\scriptsize\bfseries, green!40!black] at (0,2.6) {Execution Layer};
    \node[box, fill=white, minimum width=2.8cm] (qexec) at (-2.0,1.9)
      {QueryExecutor};
    \node[box, fill=white, minimum width=3.4cm] (cexec) at (2.0,1.9)
      {CurateExecutor + Sandbox};

    \node[layer=yellow!10, minimum width=8.0cm, minimum height=1.7cm]
      (kl) at (0,0.15) {};

    \node[draw, rounded corners=3pt, gray!50, densely dashed, semithick,
          minimum width=5.0cm, minimum height=0.8cm,
          inner sep=6pt] (store) at (-0.2,0.05) {};
    \node[box, fill=white, minimum width=2.0cm] (ct) at (-1.6,0.05)
      {Context Tree};
    \node[box, fill=white, minimum width=2.0cm] (ms) at (0.9,0.05)
      {MiniSearch};

    \node[box, fill=white, minimum width=1.2cm] (cache) at (3.1,0.05)
      {Cache};

    \node[font=\tiny, text=black!35] at (0,-0.6)
      {local filesystem --- no external services};

    \draw[darr] (llm) -- (mtools);
    \draw[arr] (llm.south) -- (qexec.north);
    \draw[arr] (mtools.south) -- (cexec.north);
    \draw[arr] (qexec.south) -- (qexec.south |- store.north);
    \draw[arr] (cexec.south) -- (cexec.south |- store.north);
    \draw[arr] ([xshift=10pt]qexec.south) -- ++(0,-0.35) -| (cache.north);

    \node[font=\scriptsize\bfseries, yellow!40!black, fill=yellow!10,
          inner sep=2pt, rounded corners=1pt] at (0,0.8) {Knowledge Layer};

  \end{tikzpicture}
  \caption{Architectural overview of \byterover{}.
    Clients (TUI, CLI, MCP) connect via Socket.IO to a daemon
    that manages a per-project task queue and agent pool.
    Each agent process contains three logical layers:
    (1)~an \emph{Agent Layer} where \texttt{curate} and
    \texttt{search\_knowledge} are first-class tools in the LLM's
    reasoning loop;
    (2)~an \emph{Execution Layer} with a query executor for 5-tier
    progressive retrieval and a sandboxed curation environment;
    and (3)~a \emph{Knowledge Layer} with the \contexttree{},
    BM25 full-text index, and query cache,
    all backed by the local filesystem with no external infrastructure.}
  \label{fig:architecture}
\end{figure}
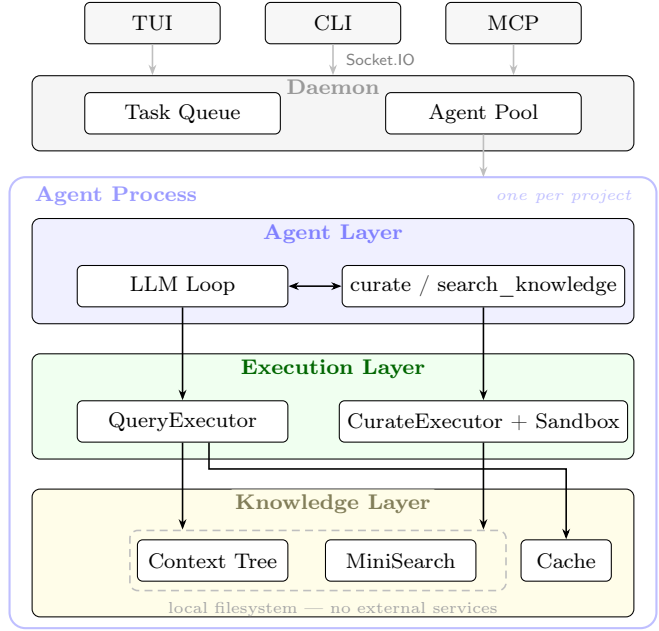

\byterover{} is organized into three logical layers,
illustrated in Figure~\ref{fig:architecture}:

\begin{itemize}[leftmargin=*]
  \item \textbf{Agent Layer.}
    The LLM reasoning loop that produces both task outputs and memory operations.
    Memory tools (curate, query, search) are available as first-class tools
    alongside file I/O, code execution, and other agent capabilities.

  \item \textbf{Execution Layer.}
    A sequential task queue that processes curate and query operations.
    Curation runs through a sandboxed environment where the LLM's generated code
    has controlled access to the knowledge layer via a \texttt{ToolsSDK} interface.
    The sequential queue eliminates write-write conflicts without file-level locking.

  \item \textbf{Knowledge Layer.}
    The \contexttree{} (a hierarchical file structure of markdown entries),
    the MiniSearch full-text index, and the query cache.
    All storage is local filesystem---no external databases or services.
\end{itemize}

The key design principle is that memory operations are \emph{tools in the agent's toolkit},
not API calls to an external service.
When the agent curates knowledge, it produces structured operations that execute within
the agent process, with per-operation feedback enabling real-time error recovery.
When the agent queries knowledge, results are returned from in-process caches and indexes
before any LLM call is considered.
Each project is served by a single agent process with its own \contexttree{},
managed by a per-project agent pool.
When multiple clients (TUI, CLI, MCP) submit tasks to the same project,
a sequential, deduplicated task queue serializes all operations,
eliminating write-write conflicts without file-level locking.
\byterover{} exposes \texttt{brv-query} and \texttt{brv-curate} as MCP tools,
enabling integration with any MCP-compatible agent framework.
All knowledge is stored as human-readable markdown files on the local filesystem---version-controllable,
portable, and requiring zero external infrastructure.

\subsection{The Context Tree: Data Structure}
\label{sec:context-tree}

The \contexttree{} is a hierarchical file-based knowledge graph
organized as \textit{Domain~$>$~Topic~$>$~Subtopic~$>$~Entry}.
We formalize it as a directed graph $\mathcal{G} = (\mathcal{N}, \mathcal{E})$
where nodes~$\mathcal{N}$ are knowledge entries (markdown files)
and edges~$\mathcal{E}$ are explicit cross-references
declared via \texttt{@domain/topic/file.md} relation annotations.

\subsubsection{Knowledge Entry Structure}

Each entry $n_i \in \mathcal{N}$ is a standalone markdown file with structured content
(Equation~\ref{eq:entry}, Appendix~\ref{app:entry-example}):
\begin{equation}
  n_i = \langle \mathcal{R}_i, \mathcal{C}_i, \mathcal{V}_i, \mathcal{S}_i, \mathcal{L}_i \rangle,
  \label{eq:entry}
\end{equation}
where $\mathcal{R}_i$ denotes the relation set (explicit edges to other entries),
$\mathcal{C}_i$ is the raw concept (provenance: task, changes, sources, timestamp, author),
$\mathcal{V}_i$ is the narrative (interpreted structure: dependencies, rules, examples, diagrams),
$\mathcal{S}_i$ contains snippets (code, formulas, raw data),
and $\mathcal{L}_i$ is the lifecycle metadata.

\subsubsection{Relation Graph and Symbol Tree}

The edge set $\mathcal{E}$ is constructed from explicit \texttt{@relation} annotations
in the \texttt{Relations} section of each entry.
Unlike embedding-based implicit similarity, these edges represent
\emph{author-stated} semantic connections---the LLM that created the entry
decided that these concepts are related and stated why.

A bidirectional reference index maintains both forward links
(source~$\to$~targets it references) and backlinks (target~$\to$~sources that reference it),
enabling graph traversal in both directions with O(1) lookup per entry.

A hierarchical \emph{symbol tree} provides O(1) lookup from relative paths to knowledge entries
and hosts the reference index above.
The tree supports five symbol kinds:
\emph{Domain}~(1), \emph{Topic}~(2), \emph{Subtopic}~(3), \emph{Context}~(4), and \emph{Summary}~(5).
For query and curate operations, a lightweight representation of the tree structure
is injected into the agent's system prompt: either a directory listing of domain
and topic names (up to 200 entries) or, when full-text search is available,
a compact instruction to use the search tool.
This gives the agent ambient awareness of what knowledge exists
without dumping full contents.

\subsubsection{Adaptive Knowledge Lifecycle (AKL)}

Each entry carries lifecycle metadata $\mathcal{L}_i$ that governs its evolution over time
through an \emph{Adaptive Knowledge Lifecycle} (AKL) mechanism:

\begin{itemize}[leftmargin=*]
  \item \textbf{Importance score} $\iota_i \in [0, 100]$:
    Tracks the value of each entry over time.
    Access events contribute a $+3$ bonus; update events contribute $+5$.
    A daily decay factor of $0.995$ prevents unbounded accumulation.

  \item \textbf{Maturity tiers}:
    Entries progress through three tiers based on importance,
    with hysteresis gaps to prevent rapid oscillation:
    \emph{draft}~$\to$~\emph{validated} (promotion at $\iota \geq 65$, demotion at $\iota < 35$; gap of 30),
    \emph{validated}~$\to$~\emph{core} (promotion at $\iota \geq 85$, demotion at $\iota < 60$; gap of 25).

  \item \textbf{Recency decay}:
    A time-dependent score $r_i = \exp(-\Delta t_i / \tau)$
    where $\Delta t_i$ is the number of days since last update
    and $\tau = 30$ is the decay constant ($\sim$21-day half-life).
\end{itemize}

The compound retrieval score (Equation~\ref{eq:score}) combines search relevance with lifecycle signals:
\begin{equation}
  \text{Score}(n_i, q) = w_r \cdot \text{BM25}(n_i, q) + w_\iota \cdot \hat{\iota}_i + w_t \cdot r_i,
  \label{eq:score}
\end{equation}
where $\hat{\iota}_i$ is the normalized importance and $w_r, w_\iota, w_t$ are tunable weights.

\section{Agent-Native Operations}
\label{sec:operations}

\subsection{LLM-Curated Knowledge Operations}
\label{sec:curation}

\byterover{} supports five atomic curate operations (Table~\ref{tab:operations}) that the LLM can compose:

\begin{table}[h]
\centering
\small
\begin{tabular}{@{}ll@{}}
\toprule
\textbf{Operation} & \textbf{Behavior} \\
\midrule
\texttt{ADD} & Create new entry; auto-generate \texttt{context.md} at each hierarchy level \\
\texttt{UPDATE} & Replace content of an existing entry \\
\texttt{UPSERT} & Add if new, update if exists (reduces pre-check overhead) \\
\texttt{MERGE} & Combine two entries intelligently; delete the source \\
\texttt{DELETE} & Remove a single entry or an entire subtree \\
\bottomrule
\end{tabular}
\caption{The five atomic curate operations.
  Every operation carries a \texttt{reason} field that serves as an audit trail.}
\label{tab:operations}
\end{table}

\subsubsection{Curation Pipeline}

Curation follows a three-phase process:

\begin{enumerate}[leftmargin=*]
  \item \textbf{Preprocessing.}
    Source documents are read and validated (max 5 files, 40K characters each).
    PDFs are converted to text; code files are truncated to 2000 lines.

  \item \textbf{Pre-Compaction.}
    An escalated compression strategy reduces input size through three levels:
    (L1)~LLM summarization,
    (L2)~aggressive LLM summarization at $0.6\times$ token budget,
    (L3)~deterministic binary-search prefix truncation (guaranteed convergence).
    This ensures curation always terminates regardless of input size.

  \item \textbf{Curation.}
    The LLM agent runs in a sandboxed environment with access to the \texttt{ToolsSDK}---a
    controlled interface providing \texttt{curate()}, \texttt{searchKnowledge()},
    \texttt{readFile()}, and other file operations.
    The agent reads sources, reasons about patterns and relationships,
    and produces structured curate operations with explicit provenance.
\end{enumerate}

\subsubsection{Stateful Feedback Loop}

A critical differentiator from external services is the \emph{stateful feedback loop}.
Each curate call returns per-operation status:
\begin{lstlisting}[firstnumber=1]
{
  "applied": [
    {"type": "UPSERT", "path": "analysis/semi", "status": "success"},
    {"type": "MERGE",  "path": "analysis/energy", "status": "failed",
     "message": "Source file not found"}
  ],
  "summary": {"added": 0, "deleted": 0, "updated": 1, "merged": 0, "failed": 1}
}
\end{lstlisting}

The agent sees which operations succeeded, which failed, and why.
It can reason about failures and adapt---skip the operation, retry with corrections,
or flag the gap for later resolution.
This feedback loop is impossible when memory is an external service
returning HTTP status codes.

\subsubsection{Atomic Writes and Crash Safety}

All file operations use an atomic write-to-temp-then-rename pattern.
If the process crashes mid-write, the \contexttree{} remains consistent---no
partial entries or corrupted knowledge.

\subsection{5-Tier Progressive Retrieval}
\label{sec:retrieval}

\begin{figure}[t]
  \centering
  \begin{tikzpicture}[
    font=\scriptsize,
    arr/.style={-{Stealth[length=3pt]}, semithick},
    step/.style={draw, rounded corners=2pt, fill=#1,
                 minimum width=4.2cm, minimum height=0.5cm,
                 text centered, font=\scriptsize},
    exitbox/.style={draw, rounded corners=2pt, fill=green!10,
                minimum width=1.4cm, minimum height=0.4cm,
                text centered, font=\tiny\bfseries},
    ylbl/.style={font=\tiny\sffamily, text=green!50!black},
    nlbl/.style={font=\tiny\sffamily, text=black!50},
  ]


    \node[step=gray!10, font=\scriptsize\bfseries] (q) at (0,0)
      {Query $q$};

    \node[step=blue!5] (par) at (0,-0.9)
      {Start search $\parallel$ Compute fingerprint};

    \node[step=blue!8] (t0) at (0,-1.8)
      {Tier 0: Exact cache match?};
    \node[exitbox] (t0y) at (3.6,-1.8) {Return};

    \node[step=blue!8] (t1) at (0,-2.7)
      {Tier 1: Fuzzy cache $\geq$0.6?};
    \node[exitbox] (t1y) at (3.6,-2.7) {Return};

    \node[step=orange!8] (aw) at (0,-3.6)
      {Await search results};

    \node[step=red!6] (ood) at (0,-4.5)
      {Results = 0? (OOD)};
    \node[exitbox, fill=red!8] (oody) at (3.6,-4.5) {Reject};

    \node[step=yellow!12] (t2) at (0,-5.4)
      {Tier 2: Score $\geq$0.85 + dominant?};
    \node[exitbox] (t2y) at (3.6,-5.4) {Return};

    \node[step=green!8] (t3) at (0,-6.3)
      {Tier 3: LLM + pre-fetched context};
    \node[exitbox] (t3y) at (3.6,-6.3) {Return};

    \node[step=red!6] (t4) at (0,-7.2)
      {Tier 4: Full agentic loop};
    \node[exitbox] (t4y) at (3.6,-7.2) {Return};

    \draw[arr] (q) -- (par);
    \draw[arr] (par) -- (t0);
    \draw[arr] (t0) -- node[nlbl, right] {miss} (t1);
    \draw[arr] (t1) -- node[nlbl, right] {miss} (aw);
    \draw[arr] (aw) -- (ood);
    \draw[arr] (ood) -- node[nlbl, right] {no} (t2);
    \draw[arr] (t2) -- node[nlbl, right] {miss} (t3);
    \draw[arr] (t3) -- node[nlbl, right] {no ctx} (t4);

    \draw[arr] (t0) -- node[ylbl, above] {hit} (t0y);
    \draw[arr] (t1) -- node[ylbl, above] {hit} (t1y);
    \draw[arr] (ood) -- node[ylbl, above] {yes} (oody);
    \draw[arr] (t2) -- node[ylbl, above] {hit} (t2y);
    \draw[arr] (t3) -- (t3y);
    \draw[arr] (t4) -- (t4y);

    \node[font=\tiny, text=black!35, anchor=west] at (4.5,-1.8) {0\,ms};
    \node[font=\tiny, text=black!35, anchor=west] at (4.5,-2.7) {$\sim$50\,ms};
    \node[font=\tiny, text=black!35, anchor=west] at (4.5,-5.4) {$\sim$200\,ms};
    \node[font=\tiny, text=black!35, anchor=west] at (4.5,-6.3) {$<$5\,s};
    \node[font=\tiny, text=black!35, anchor=west] at (4.5,-7.2) {8--15\,s};

    \node[font=\tiny, text=blue!50, anchor=east] at (-2.5,-1.8) {Cache};
    \node[font=\tiny, text=blue!50, anchor=east] at (-2.5,-2.7) {Cache};
    \node[font=\tiny, text=yellow!50!black, anchor=east] at (-2.5,-5.4) {MiniSearch};
    \node[font=\tiny, text=green!50!black, anchor=east] at (-2.5,-6.3) {LLM};
    \node[font=\tiny, text=red!50!black, anchor=east] at (-2.5,-7.2) {Agent};

    \draw[decorate, decoration={brace, amplitude=3pt, mirror},
          thick, gray!40]
      (-4.0,-0.65) -- (-4.0,-5.65)
      node[midway, left=5pt, font=\tiny, text=black!40, align=right]
        {no LLM\\call};

  \end{tikzpicture}
  \caption{The 5-tier progressive retrieval pipeline.
    Search is initiated in parallel with fingerprint computation.
    Tiers~0--1 resolve from cache without awaiting search.
    Tier~2 serves high-confidence results directly from MiniSearch.
    Only novel or ambiguous queries escalate to
    Tier~3 (single optimized LLM call with pre-fetched context) or
    Tier~4 (full agentic loop with tool access).
    Approximate latencies shown on right.}
  \label{fig:retrieval}
\end{figure}
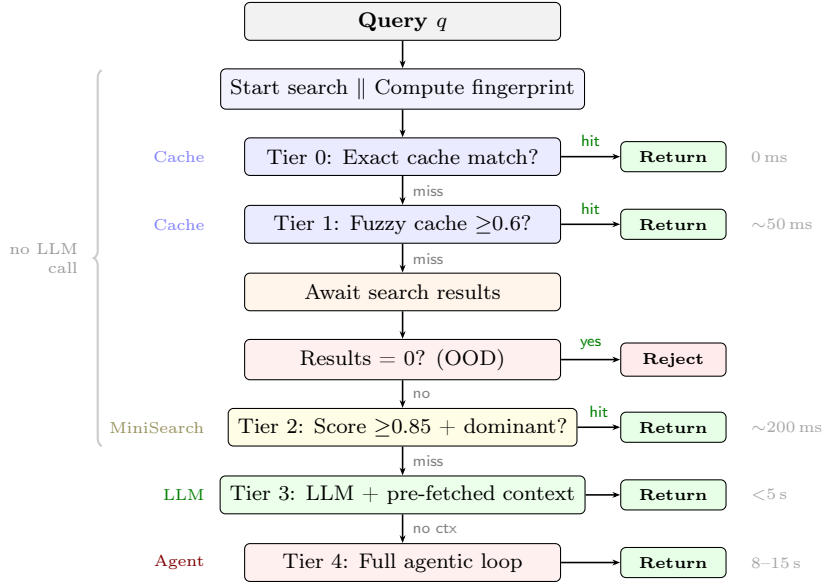

Retrieval uses a tiered strategy (Table~\ref{tab:tiers}) that minimizes LLM calls,
illustrated in Figure~\ref{fig:retrieval} and formalized in Algorithm~\ref{alg:retrieval}.

\begin{algorithm}[t]
\caption{5-Tier Progressive Retrieval}
\label{alg:retrieval}
\begin{algorithmic}[1]
\REQUIRE Query $q$, Cache $\mathcal{C}$, Search Index $\mathcal{I}$, Agent $\mathcal{A}$
\ENSURE Response $R$, Tier $\tau$
\STATE $h \leftarrow \text{Hash}(q)$
\IF{$h \in \mathcal{C}$ \AND $\text{Fingerprint}(\mathcal{C}[h]) = \text{Fingerprint}(\mathcal{G})$}
  \RETURN $\mathcal{C}[h]$, $\tau = 0$ \COMMENT{Exact cache hit}
\ENDIF
\STATE $q' \leftarrow \arg\max_{c \in \mathcal{C}} \text{Jaccard}(q, c)$
\IF{$\text{Jaccard}(q, q') \geq \theta_{\text{fuzzy}}$}
  \RETURN $\mathcal{C}[q']$, $\tau = 1$ \COMMENT{Fuzzy cache hit}
\ENDIF
\STATE $\mathcal{D} \leftarrow \text{MiniSearch}(\mathcal{I}, q)$ \COMMENT{BM25 + fuzzy + prefix}
\IF{$\text{Score}(\mathcal{D}_1) \geq \theta_{\text{high}}$ \AND $\text{Gap}(\mathcal{D}_1, \mathcal{D}_2) \geq \theta_{\text{gap}}$}
  \RETURN $\text{DirectResponse}(\mathcal{D})$, $\tau = 2$ \COMMENT{High-confidence search}
\ENDIF
\IF{$\text{Score}(\mathcal{D}_1) \geq \theta_{\text{med}}$}
  \STATE $C_{\text{pre}} \leftarrow \text{Prefetch}(\mathcal{D})$
  \RETURN $\text{LLM}(q, C_{\text{pre}})$, $\tau = 3$ \COMMENT{Optimized single LLM call}
\ENDIF
\RETURN $\mathcal{A}.\text{AgenticLoop}(q)$, $\tau = 4$ \COMMENT{Full multi-turn reasoning}
\end{algorithmic}
\end{algorithm}

\begin{table}[t]
\centering
\small
\begin{tabular}{@{}clcl@{}}
\toprule
\textbf{Tier} & \textbf{Mechanism} & \textbf{Latency} & \textbf{Condition} \\
\midrule
0 & Exact cache hit & $\sim$0\,ms & Hash match + valid fingerprint \\
1 & Fuzzy cache (Jaccard) & $\sim$50\,ms & Jaccard $\geq \theta_{\text{fuzzy}}$ \\
2 & Direct MiniSearch & $\sim$100\,ms & BM25 score $\geq \theta_{\text{high}}$, sufficient gap \\
3 & Optimized LLM call & $<$5\,s & BM25 score $\geq \theta_{\text{med}}$ \\
4 & Full agentic loop & 8--15\,s & All other queries \\
\bottomrule
\end{tabular}
\caption{The five retrieval tiers with their latency characteristics and escalation conditions.}
\label{tab:tiers}
\end{table}

Tiers~0--2 resolve queries without any LLM call, returning cached
or high-confidence search results directly.
Tier~3 pre-fetches relevant documents and passes them
to a single optimized LLM call with constrained output
(1{,}024 tokens, temperature~0.3).
Tier~4 is the full agentic fallback:
the agent enters a multi-turn reasoning loop
where it calls tools (\texttt{code\_exec}, \texttt{readFile})
to navigate the \contexttree{}, with a higher token budget
(2{,}048 tokens, temperature~0.5) and up to 50 iterations.

\subsubsection{Search Engine}

The search engine is MiniSearch---a lightweight full-text search library
with BM25 ranking, fuzzy matching (0.2 character similarity threshold),
and prefix search.
Field boosting weights titles at $5\times$ and paths at $1.5\times$ over content.

Score normalization (Equation~\ref{eq:score-norm}) maps raw BM25 scores to $[0, 1)$ via:
\begin{equation}
  \hat{s} = \frac{s_{\text{raw}}}{1 + s_{\text{raw}}},
  \label{eq:score-norm}
\end{equation}
yielding interpretable thresholds: strong (15)~$\to$~0.94, medium (8)~$\to$~0.89,
moderate (4)~$\to$~0.80, weak (1)~$\to$~0.50.

\subsubsection{Out-of-Domain Detection}

When significant query terms (length~$\geq$~4 characters) do not match
any entry in the knowledge base and the normalized score falls below
a threshold ($\theta_{\text{OOD}} = 0.85$), the system explicitly signals
\emph{``this query appears outside the scope of stored knowledge.''}
This prevents hallucinated answers from tangential results---an
essential property when agents are making decisions based on retrieved knowledge.

\section{Experiments}
\label{sec:experiments}

We conduct comprehensive experiments to evaluate both the reasoning effectiveness
and system properties of \byterover{} over state-of-the-art baselines
(Tables~\ref{tab:locomo}--\ref{tab:efficiency}).

\subsection{Experimental Setup}
\label{sec:exp-setup}

\paragraph{Datasets.}
We evaluate on two widely adopted long-term conversational benchmarks:
\textbf{(1)~LoCoMo}~\citep{maharana2024locomo},
which contains ultra-long conversations (average length of $\sim$20K tokens across 35 sessions)
designed to assess long-range temporal and causal retrieval;
and \textbf{(2)~LongMemEval-S}~\citep{wu2024longmemeval},
the full-scale variant of LongMemEval with 500 questions spanning six memory-ability categories,
average context length exceeding 100K tokens across ${\sim}$48 sessions per question,
designed to stress-test memory retention and scalability under realistic interaction horizons.

\paragraph{Baselines.}
On LoCoMo, we evaluate six memory systems under a unified protocol
using our benchmark harness with identical judge configuration,
enabling direct comparison:
\textbf{Mem0}~\citep{chhikara2025mem0},
\textbf{Zep}~\citep{rasmussen2025zep},
\textbf{Hindsight}~\citep{latimer2025hindsight},
\textbf{HonCho},\footnote{\url{https://honcho.dev}}
\textbf{Memobase},\footnote{\url{https://github.com/memodb-io/memobase}}
and \textbf{OpenAI Memory} (ChatGPT).
These systems span the spectrum from structured fact extraction
(Mem0, Memobase) to graph-based memory architectures (Zep, Hindsight)
and commercial assistants (OpenAI Memory, HonCho).

On LongMemEval-S, we include our own evaluations of Hindsight and HonCho
alongside published results from
\textbf{Chronos}~\citep{sen2026chronos},
\textbf{SmartSearch}~\citep{derehag2026smartsearch},
\textbf{Memora}~\citep{xia2026memora},
\textbf{TiMem}~\citep{li2026timem},
Zep, and a full-context baseline,
cited from their respective publications.
Results marked with $\dagger$ use different backbone and judge configurations;
see Table~\ref{tab:longmemeval} for details.

\paragraph{Metrics.}
We adopt LLM-as-a-Judge~\citep{zheng2023llmjudge} as the primary metric,
where a judge model evaluates each generated answer against the ground truth
and returns a binary correctness label.
For our own evaluations, we use Gemini~3 Flash as the judge
with a separate justifier model (Gemini~3.1 Pro) that synthesizes
an answer from retrieved context before scoring.
Both models run at temperature~0.0.
The judge uses a token budget of 8{,}192 with thinking disabled,
while the justifier uses 32{,}768 tokens with low thinking budget.
Full hyperparameter details are provided in Appendix~\ref{app:hyperparameters}.

\paragraph{Evaluation prompts.}
To enable direct, apples-to-apples comparison with Hindsight~\citep{latimer2025hindsight},
we reuse their publicly available evaluation prompts.\footnote{%
  \url{https://github.com/vectorize-io/hindsight/tree/main/hindsight-dev/benchmarks}}
Specifically, the judge prompts---including the default LoCoMo preamble
and all five LongMemEval category-specific preambles
(standard, temporal-reasoning, knowledge-update, and preference)---are
adopted without modification, ensuring that scoring criteria
are identical across systems.
The LoCoMo justifier prompt is likewise reused verbatim.
For the LongMemEval justifier, we adapt Hindsight's prompt
to reflect \byterover{}'s single-layer knowledge representation:
where Hindsight instructs the model to cross-reference
atomic facts against raw source chunks (a two-layer retrieval output),
our variant replaces this with guidance for interpreting
\emph{key facts} with session-level timestamps,
matching the \contexttree{}'s curated entry format.
All other instructions---date calculation rules, counting-question
disambiguation, recommendation handling, and question-type-specific
guidelines---remain unchanged from the original.

\subsection{Overall Comparison on LoCoMo}
\label{sec:locomo-results}

\begin{table}[t]
\centering
\small
\begin{tabular}{@{}lcccc|c@{}}
\toprule
\textbf{Method} & \textbf{Single-Hop} & \textbf{Multi-Hop} & \textbf{Open-Domain} & \textbf{Temporal} & \textbf{Overall} \\
\midrule
HonCho          & 93.2 & 84.0 & 77.1 & 88.2 & 89.9 \\
Hindsight       & 86.2 & 70.8 & \textbf{95.1} & 83.8 & 89.6 \\
Memobase        & 70.9 & 46.9 & 77.2 & 85.1 & 75.8 \\
Zep             & 74.1 & 66.0 & 67.7 & 79.8 & 75.1 \\
Mem0            & 67.1 & 51.2 & 72.9 & 55.5 & 66.9 \\
OpenAI Memory   & 63.8 & 42.9 & 62.3 & 21.7 & 52.9 \\
\midrule
\textbf{\byterover{} (ours)} & \textbf{97.5} & \textbf{93.3} & 85.9 & \textbf{97.8} & \textbf{96.1} \\
\bottomrule
\end{tabular}
\caption{LLM-as-Judge accuracy (\%) on LoCoMo (4 categories, 1{,}982 questions).
  All systems evaluated under identical conditions using our benchmark harness
  with Gemini~3 Flash as the judge.}
\label{tab:locomo}
\end{table}

\byterover{} achieves the highest overall accuracy at 96.1\%,
outperforming the next-best system (HonCho, 89.9\%) by 6.2 percentage points.
The gains are particularly pronounced on \emph{multi-hop} questions
($+$9.3 over HonCho),
which require synthesizing information across distant sessions---a
task where the \contexttree{}'s explicit inter-entry relations
provide navigable paths that flat memory stores lack.
On \emph{temporal} queries, \byterover{} reaches 97.8\%,
benefiting from structured timestamps embedded in each entry's metadata
that enable precise temporal grounding.
The sole category where \byterover{} does not lead is \emph{open-domain},
where Hindsight achieves 95.1\% versus 85.9\%.
Open-domain questions frequently require commonsense reasoning
that extends beyond the conversation corpus,
an area where retrieval-augmented approaches can excel
by leveraging the backbone LLM's parametric knowledge.

\subsection{Generalization on LongMemEval-S}
\label{sec:longmemeval-results}

\begin{table}[t]
\centering
\small
\begin{tabular}{@{}lcccccc|c@{}}
\toprule
\textbf{Method} & \textbf{KU} & \textbf{SSU} & \textbf{SSA} & \textbf{SSP} & \textbf{TR} & \textbf{MS} & \textbf{Overall} \\
\midrule
Chronos$^\dagger$       & 96.2 & 94.3 & \textbf{100.0} & 80.0 & 90.2 & \textbf{91.7} & 92.6 \\
Hindsight               & 94.9 & 97.1 & 96.4 & 80.0 & 91.0 & 87.2 & 91.4 \\
HonCho                  & 94.9 & 94.3 & 96.4 & 90.0 & 88.7 & 85.0 & 90.4 \\
SmartSearch$^\dagger$   & 93.6 & \textbf{100.0} & 85.7 & \textbf{96.7} & 82.7 & 84.2 & 88.4 \\
Memora$^\dagger$        & 97.4 & 98.6 & 78.6 & 83.3 & 89.5 & 78.2 & 87.4 \\
TiMem$^\dagger$         & 87.7 & 96.3 & 85.7 & 55.3 & 73.4 & 72.8 & 79.0 \\
Zep$^\dagger$           & 83.3 & 92.9 & 80.4 & 56.7 & 62.4 & 57.9 & 71.2 \\
Full-context$^\dagger$  & 78.2 & 81.4 & 94.6 & 20.0 & 45.1 & 44.3 & 60.2 \\
\midrule
\textbf{\byterover{} (ours)} & \textbf{98.7} & 98.6 & 98.2 & \textbf{96.7} & \textbf{91.7} & 84.2 & \textbf{92.8} \\
\bottomrule
\end{tabular}
\caption{LLM-as-Judge accuracy (\%) on LongMemEval-S (500 questions).
  KU\,=\,Knowledge Update, SSU\,=\,Single-Session User, SSA\,=\,Single-Session Assistant,
  SSP\,=\,Single-Session Preference, TR\,=\,Temporal Reasoning, MS\,=\,Multi-Session.
  Hindsight and HonCho evaluated with our harness (Gemini~3 Flash judge).
  $^\dagger$Results from respective papers; backbone and judge configurations differ.}
\label{tab:longmemeval}
\end{table}

\byterover{} achieves the highest overall accuracy at 92.8\%,
surpassing Chronos-Low (92.6\%, GPT-4o backbone)
while remaining below Chronos-High (95.6\%, Claude Opus~4.6 backbone),
which benefits from a substantially more capable generation model.
The strength profile is distinctive:
\byterover{} leads on knowledge update (98.7\%),
temporal reasoning (91.7\%), and single-session preference
(96.7\%, tied with SmartSearch),
categories that reward precise fact retrieval and temporal grounding---capabilities
directly served by the \contexttree{}'s structured entries
and timestamp-aware indexing.
The weakest category is multi-session (84.2\%),
which requires cross-session synthesis over long horizons
and represents the primary area for improvement.
Notably, Chronos achieves 91.7\% on multi-session,
suggesting that its event-ordering approach
may better capture inter-session dependencies.

\subsection{Operational Profile}
\label{sec:efficiency}

\begin{table}[t]
\centering
\small
\begin{tabular}{@{}lcc@{}}
\toprule
\textbf{Metric} & \textbf{LoCoMo} & \textbf{LongMemEval-S} \\
\midrule
Context tree size & 272 docs & 23{,}867 docs \\
Total queries & 1{,}982 & 500 \\
\midrule
Cold query latency p50 & 1.2\,s & 1.6\,s \\
Cold query latency p95 & 1.4\,s & 2.3\,s \\
Cold query latency p99 & 1.7\,s & 2.5\,s \\
\bottomrule
\end{tabular}
\caption{Operational latency profile.
  Cold query latency measures end-to-end time from process invocation
  through response delivery, excluding answer justification and evaluation.}
\label{tab:efficiency}
\end{table}

Despite a substantially larger context tree on LongMemEval-S
(23{,}867 documents vs.\ 272 on LoCoMo),
median query latency remains low at 1.6\,s,
suggesting that the tiered retrieval architecture
effectively bounds search cost as the corpus grows.
The tight percentile spread (p50-to-p99 gap of 0.5\,s on LoCoMo,
0.9\,s on LongMemEval-S) indicates consistent performance
without long-tail degradation.

\subsection{Ablation Study}
\label{sec:ablation}

To assess the contribution of individual components,
we conduct an ablation study on LongMemEval-S by selectively disabling
three key query-time mechanisms.
Ablation experiments are conducted on LongMemEval-S,
where the larger corpus (23{,}867 documents) provides a more demanding
test of retrieval-side components;
LoCoMo results (Table~\ref{tab:locomo}) serve as a generalization
test on a smaller corpus.

\begin{table}[t]
\centering
\small
\begin{tabular}{@{}lcccccc|cc@{}}
\toprule
\textbf{Configuration} & \textbf{KU} & \textbf{SSU} & \textbf{SSA} & \textbf{SSP} & \textbf{TR} & \textbf{MS} & \textbf{Overall} & \textbf{$\Delta$} \\
\midrule
w/o Tiered Retrieval   & 69.2 & 71.4 & 76.8 & 83.3 & 61.7 & 47.4 & 63.4 & $-$29.4 \\
w/o OOD Detection      & 98.7 & 98.6 & 98.2 & 96.7 & 89.5 & 85.0 & 92.4 & $-$0.4 \\
w/o Relation Graph     & 98.7 & 98.6 & 98.2 & 96.7 & 89.5 & 85.0 & 92.4 & $-$0.4 \\
\midrule
\textbf{\byterover{} (Full)} & 98.7 & 98.6 & 98.2 & 96.7 & 91.7 & 84.2 & \textbf{92.8} & --- \\
\bottomrule
\end{tabular}
\caption{Ablation study on LongMemEval-S (500 questions).
  Each row disables one query-time component while keeping the curated
  \contexttree{} unchanged.
  Category abbreviations follow Table~\ref{tab:longmemeval}.}
\label{tab:ablation}
\end{table}

\paragraph{w/o Tiered Retrieval.}
All queries are routed to the full agentic loop (Tier~4),
bypassing Tiers~0--3 and the separate justifier model
(see~\S\ref{sec:retrieval} for the distinction between tiers).
This ablation produces the largest accuracy drop ($-$29.4~pp, from 92.8\% to 63.4\%).
Multi-session questions suffer most severely (84.2\%~$\to$~47.4\%),
followed by temporal reasoning (91.7\%~$\to$~61.7\%).
The result demonstrates that the tiered architecture is not merely
a latency optimization: the lower tiers surface precise,
high-confidence content that the justifier can faithfully synthesize,
whereas unconstrained agentic reasoning over a 23{,}867-document
corpus compounds retrieval errors with generation errors.

\paragraph{w/o OOD Detection.}
Disabling the out-of-domain rejection gate produces
only a modest overall decline ($-$0.4~pp, from 92.8\% to 92.4\%),
with the effect concentrated in temporal reasoning
(91.7\%~$\to$~89.5\%, $-$2.2~pp).
Four categories (KU, SSU, SSA, SSP) are completely unaffected,
indicating that OOD detection primarily guards against
temporal and multi-session queries where partial matches
from unrelated sessions could otherwise mislead the justifier.
The small magnitude suggests the curated \contexttree{}
already provides strong topical coherence,
limiting the opportunities for OOD errors.

\paragraph{w/o Relation Graph.}
Removing explicit cross-entry relations produces the same
overall decline as disabling OOD detection ($-$0.4~pp, 92.4\%),
with an identical per-category profile:
temporal reasoning drops by 2.2~pp (91.7\%~$\to$~89.5\%)
while multi-session improves slightly (84.2\%~$\to$~85.0\%).
The identical scores to the OOD ablation suggest that,
on LongMemEval-S's question distribution,
relation edges and OOD gates address overlapping failure modes---both
primarily affect queries that require disambiguating
temporally adjacent sessions.
The relation graph's contribution may be more pronounced
on benchmarks with explicit multi-hop reasoning demands
(e.g., LoCoMo's multi-hop category, where relations
provide navigable paths across conversation boundaries).

\paragraph{Components not ablated.}
Three mechanisms---Adaptive Knowledge Lifecycle (AKL), the curation feedback loop,
and escalated compression---operate exclusively during the write path (curation).
Since the curated \contexttree{} is held constant across all ablation conditions,
their contribution cannot be isolated through query-time ablation on a static benchmark.
Evaluating these components would require re-curation under degraded conditions,
which we leave to future work.

\section{Conclusion}
\label{sec:conclusion}

We introduced \byterover{}, an agent-native memory architecture that inverts
the conventional MAG paradigm: instead of delegating knowledge storage to an external service,
the same LLM that reasons about a task curates knowledge into a hierarchical \contexttree{}
with explicit relations, provenance, and lifecycle management.
By making memory operations first-class tools in the agent's reasoning loop,
\byterover{} enables a stateful feedback loop that eliminates semantic drift,
preserves coordination context across agents, and supports graceful recovery from failures.

The 5-tier progressive retrieval strategy resolves most queries at sub-100\,ms latency
without LLM calls, while the Adaptive Knowledge Lifecycle enables knowledge
to naturally evolve---reinforcing frequently accessed entries and decaying stale ones.
Empirical results on LoCoMo and LongMemEval demonstrate that \byterover{}
achieves competitive or state-of-the-art accuracy while requiring
zero external infrastructure, with all knowledge stored as human-readable markdown files
on the local filesystem.

\section{Limitations}
\label{sec:limitations}

While \byterover{} demonstrates strong architectural properties,
it has several limitations that warrant discussion.

First, the \textbf{write path is expensive}.
LLM-curated knowledge requires reasoning per curation event,
which is slower and more costly than mechanical chunking and embedding.
For use cases where write throughput is critical
(e.g., real-time ingestion of high-frequency data streams),
the curation overhead may be prohibitive.

Second, \textbf{novel queries are slower than vector search}.
When queries miss the cache and index (Tier~3--4),
\byterover{} requires an LLM call that a vector similarity search does not.
The design assumes that agents ask many variations of a smaller set of questions,
so the cache absorbs most of the load.
When this assumption does not hold, retrieval latency increases.

Third, \textbf{curation quality depends on backbone model capability}.
This is a shared limitation across all MAG systems~\citep{jiang2026anatomy},
but \byterover{}'s deeper reliance on LLM reasoning for both storage and retrieval
amplifies the impact of backbone sensitivity.
Open-weight models with higher format error rates may produce
lower-quality knowledge entries.

Fourth, the \textbf{file-based storage may face scaling challenges} at very large knowledge bases.
The in-memory MiniSearch index and sequential task queue are designed for
knowledge bases of up to $\sim$10K entries.
Beyond this scale, sharding strategies or alternative indexing backends may be needed.

Finally, the \textbf{sequential task queue limits write throughput}
under heavy concurrent load.
While this is rarely a bottleneck in practice
(curation is an infrequent, high-cost operation),
deployments with many agents writing simultaneously may experience queuing delays.

\bibliographystyle{plainnat}
\bibliography{references}

@article{jiang2026magma,
  title={MAGMA: A Multi-Graph based Agentic Memory Architecture for AI Agents},
  author={Jiang, Dongming and Li, Yi and Li, Guanpeng and Li, Bingzhe},
  journal={arXiv preprint arXiv:2601.03236},
  year={2026}
}

@inproceedings{xu2025amem,
  title={{A-MEM}: Agentic Memory for LLM Agents},
  author={Xu, Wujiang and Liang, Zujie and Mei, Kai and Gao, Hang and Tan, Juntao and Zhang, Yongfeng},
  booktitle={Advances in Neural Information Processing Systems},
  volume={38},
  year={2025}
}

@article{kang2025memoryos,
  title={Memory OS of AI Agent},
  author={Kang, Jiazheng and Ji, Mingming and Zhao, Zhe and Bai, Ting},
  journal={arXiv preprint arXiv:2506.06326},
  year={2025}
}

@article{nan2025nemori,
  title={Nemori: Self-Organizing Agent Memory Inspired by Cognitive Science},
  author={Nan, Jiayan and Ma, Wenquan and Wu, Wenlong and Chen, Yize},
  journal={arXiv preprint arXiv:2508.03341},
  year={2025}
}

@article{rasmussen2025zep,
  title={Zep: A Temporal Knowledge Graph Architecture for Agent Memory},
  author={Rasmussen, Preston and Paliychuk, Pavlo and Beauvais, Travis and Ryan, Jack and Chalef, Daniel},
  journal={arXiv preprint arXiv:2501.13956},
  year={2025}
}

@article{liu2026simplemem,
  title={SimpleMem: Efficient Lifelong Memory for LLM Agents},
  author={Liu, Jiaqi and Su, Yaofeng and Xia, Peng and Han, Siwei and Zheng, Zeyu and Xie, Cihang and Ding, Mingyu and Yao, Huaxiu},
  journal={arXiv preprint arXiv:2601.02553},
  year={2026}
}

@article{latimer2025hindsight,
  title={Hindsight is 20/20: Building Agent Memory that Retains, Recalls, and Reflects},
  author={Latimer, Chris and Boschi, Nicolo and Neeser, Andrew and Bartholomew, Chris and Srivastava, Gaurav and Wang, Xuan and Ramakrishnan, Naren},
  journal={arXiv preprint arXiv:2512.12818},
  year={2025}
}

@article{sen2026chronos,
  title={Chronos: Temporal-Aware Conversational Agents with Structured Event Retrieval for Long-Term Memory},
  author={Sen, Sahil and Lumer, Elias and Gulati, Anmol and Subbiah, Vamse Kumar},
  journal={arXiv preprint arXiv:2603.16862},
  year={2026}
}

@article{derehag2026smartsearch,
  title={SmartSearch: How Ranking Beats Structure for Conversational Memory Retrieval},
  author={Derehag, Jesper and Calva, Carlos and Ghiurau, Timmy},
  journal={arXiv preprint arXiv:2603.15599},
  year={2026}
}

@article{xia2026memora,
  title={Memora: A Harmonic Memory Representation Balancing Abstraction and Specificity},
  author={Xia, Menglin and Zhang, Xuchao and Dixit, Shantanu and Harimurugan, Paramaguru and Wang, Rujia and Ruhle, Victor and Sim, Robert and Bansal, Chetan and Rajmohan, Saravan},
  journal={arXiv preprint arXiv:2602.03315},
  year={2026}
}

@article{jiang2026anatomy,
  title={Anatomy of Agentic Memory: Taxonomy and Empirical Analysis of Evaluation and System Limitations},
  author={Jiang, Dongming and Li, Yi and Wei, Songtao and Yang, Jinxin and Kishore, Ayushi and Zhao, Alysa and Kang, Dingyi and Hu, Xu and Chen, Feng and Li, Qiannan and Li, Bingzhe},
  journal={arXiv preprint arXiv:2602.19320},
  year={2026}
}

@article{brown2020language,
  title={Language Models are Few-Shot Learners},
  author={Brown, Tom and Mann, Benjamin and Ryder, Nick and Subbiah, Melanie and Kaplan, Jared D and Dhariwal, Prafulla and Neelakantan, Arvind and Shyam, Pranav and Sastry, Girish and Askell, Amanda and others},
  journal={Advances in Neural Information Processing Systems},
  volume={33},
  pages={1877--1901},
  year={2020}
}

@article{achiam2023gpt4,
  title={GPT-4 Technical Report},
  author={Achiam, Josh and Adler, Steven and Agarwal, Sandhini and Ahmad, Lama and Akkaya, Ilge and Aleman, Florencia Leoni and Almeida, Diogo and Altenschmidt, Janko and Altman, Sam and Anadkat, Shyamal and others},
  journal={arXiv preprint arXiv:2303.08774},
  year={2023}
}

@article{wei2022chain,
  title={Chain-of-Thought Prompting Elicits Reasoning in Large Language Models},
  author={Wei, Jason and Wang, Xuezhi and Schuurmans, Dale and Bosma, Maarten and Xia, Fei and Chi, Ed and Le, Quoc V and Zhou, Denny},
  journal={Advances in Neural Information Processing Systems},
  volume={35},
  pages={24824--24837},
  year={2022}
}

@article{beltagy2020longformer,
  title={Longformer: The Long-Document Transformer},
  author={Beltagy, Iz and Peters, Matthew E and Cohan, Arman},
  journal={arXiv preprint arXiv:2004.05150},
  year={2020}
}

@inproceedings{press2021train,
  title={Train Short, Test Long: Attention with Linear Biases Enables Input Length Extrapolation},
  author={Press, Ofir and Smith, Noah A and Lewis, Mike},
  booktitle={International Conference on Learning Representations},
  year={2022}
}

@article{liu2024lost,
  title={Lost in the Middle: How Language Models Use Long Contexts},
  author={Liu, Nelson F and Lin, Kevin and Hewitt, John and Paranjape, Ashwin and Bevilacqua, Michele and Petroni, Fabio and Liang, Percy},
  journal={Transactions of the Association for Computational Linguistics},
  volume={12},
  pages={157--173},
  year={2024}
}

@article{lewis2020rag,
  title={Retrieval-Augmented Generation for Knowledge-Intensive {NLP} Tasks},
  author={Lewis, Patrick and Perez, Ethan and Piktus, Aleksandra and Petroni, Fabio and Karpukhin, Vladimir and Goyal, Naman and K{\"u}ttler, Heinrich and Lewis, Mike and Yih, Wen-tau and Rockt{\"a}schel, Tim and others},
  journal={Advances in Neural Information Processing Systems},
  volume={33},
  pages={9459--9474},
  year={2020}
}

@inproceedings{wang2024mrag,
  title={M-RAG: Reinforcing Large Language Model Performance Through Retrieval-Augmented Generation with Multiple Partitions},
  author={Wang, Zheng and Teo, Shu and Ouyang, Jieer and Xu, Yongjun and Shi, Wei},
  booktitle={Proceedings of the 62nd Annual Meeting of the Association for Computational Linguistics (Volume 1: Long Papers)},
  pages={1966--1978},
  year={2024}
}

@article{jiang2024longrag,
  title={LongRAG: Enhancing Retrieval-Augmented Generation with Long-Context LLMs},
  author={Jiang, Ziyan and Ma, Xueguang and Chen, Wenhu},
  journal={arXiv preprint arXiv:2406.15319},
  year={2024}
}

@inproceedings{zhong2024memorybank,
  title={MemoryBank: Enhancing Large Language Models with Long-Term Memory},
  author={Zhong, Wanjun and Guo, Lianghong and Gao, Qiqi and Ye, He and Wang, Yanlin},
  booktitle={Proceedings of the AAAI Conference on Artificial Intelligence},
  volume={38},
  pages={19724--19731},
  year={2024}
}

@inproceedings{park2023generative,
  title={Generative Agents: Interactive Simulacra of Human Behavior},
  author={Park, Joon Sung and O'Brien, Joseph C and Cai, Carrie Jun and Ringel Morris, Meredith and Liang, Percy and Bernstein, Michael S},
  booktitle={Proceedings of the 36th Annual ACM Symposium on User Interface Software and Technology},
  pages={1--22},
  year={2023}
}

@article{packer2023memgpt,
  title={MemGPT: Towards LLMs as Operating Systems},
  author={Packer, Charles and Fang, Vivian and Patil, Shishir G and Lin, Kevin and Wooders, Sarah and Gonzalez, Joseph E},
  journal={arXiv preprint arXiv:2310.08560},
  year={2023}
}

@article{chhikara2025mem0,
  title={Mem0: Building Production-Ready AI Agents with Scalable Long-Term Memory},
  author={Chhikara, Prateek and Khant, Dev and Aryan, Saket and Singh, Taranjeet and Yadav, Deshraj},
  journal={arXiv preprint arXiv:2504.19413},
  year={2025}
}

@article{hu2026evermemos,
  title={EverMemOS: A Self-Organizing Memory Operating System for Structured Long-Horizon Reasoning},
  author={Hu, Chuanrui and Gao, Xingze and Zhou, Zuyi and Xu, Dannong and Bai, Yi and Li, Xintong and Zhang, Hui and Li, Tong and Zhang, Chong and Bing, Lidong and others},
  journal={arXiv preprint arXiv:2601.02163},
  year={2026}
}

@article{maharana2024locomo,
  title={Evaluating Very Long-Term Conversational Memory of LLM Agents},
  author={Maharana, Adyasha and Lee, Dong-Ho and Tulyakov, Sergey and Bansal, Mohit and Barbieri, Francesco and Fang, Yuwei},
  journal={arXiv preprint arXiv:2402.17753},
  year={2024}
}

@inproceedings{wu2024longmemeval,
  title={LongMemEval: Benchmarking Chat Assistants on Long-Term Interactive Memory},
  author={Wu, Di and Wang, Hongwei and Yu, Wenhao and Zhang, Yuwei and Chang, Kai-Wei and Yu, Dong},
  booktitle={International Conference on Learning Representations},
  year={2025}
}

@article{zheng2023llmjudge,
  title={Judging LLM-as-a-Judge with MT-Bench and Chatbot Arena},
  author={Zheng, Lianmin and Chiang, Wei-Lin and Sheng, Ying and Zhuang, Siyuan and Wu, Zhanghao and Zhuang, Yonghao and Lin, Zi and Li, Zhuohan and Li, Dacheng and Xing, Eric and others},
  journal={Advances in Neural Information Processing Systems},
  volume={36},
  pages={46595--46623},
  year={2023}
}

@article{kang2025lm2,
  title={LM2: Large Memory Models},
  author={Kang, Jikun and Wu, Wenqi and Christianos, Filippos and Chan, Alex James and Greenlee, Fraser David and Thomas, George and Purtorab, Marvin and Toulis, Andrew},
  journal={arXiv preprint arXiv:2502.06049},
  year={2025}
}

@article{li2026timem,
  title={TiMem: Temporal-Hierarchical Memory Consolidation for Long-Horizon Conversational Agents},
  author={Li, Kai and Yu, Xuanqing and Ni, Ziyi and Zeng, Yi and Xu, Yao and Zhang, Xin and Li, Jitao and Sang, Xiaogang and Duan, Xuelei and Wang, Xiaogang and others},
  journal={arXiv preprint arXiv:2601.02845},
  year={2026}
}

@article{modarressi2023retllm,
  title={{RET-LLM}: Towards a General Read-Write Memory for Large Language Models},
  author={Modarressi, Ali and Imani, Ayyoob and Fayyaz, Mohsen and Sch{\"u}tze, Hinrich},
  journal={arXiv preprint arXiv:2305.14322},
  year={2023}
}

@inproceedings{gutierrez2025rag2memory,
  title={From RAG to Memory: Non-Parametric Continual Learning for Large Language Models},
  author={Guti{\'e}rrez, Bernal Jim{\'e}nez and Shu, Yiheng and Qi, Sizhe and Zhou, Weijian and Su, Yu},
  booktitle={International Conference on Machine Learning},
  year={2025}
}

\clearpage
\appendix

\section{Related Work}
\label{app:related-work}

We organize related work along the progression from context-window extension
to retrieval-augmented generation and finally to memory-augmented generation,
then discuss the external-service paradigm that \byterover{} departs from.

\paragraph{Context-Window Extension.}
A direct line of work extends the effective context length of Transformers
by modifying attention or positional extrapolation.
Longformer~\citep{beltagy2020longformer} introduces sparse attention patterns,
while ALiBi~\citep{press2021train} enables length extrapolation
by injecting distance-aware linear biases into attention scores.
More recently, LM2~\citep{kang2025lm2} proposes a decoder-only architecture
augmented with auxiliary memory.
While these approaches improve long-range coverage,
they do not address the continual, evolving, and write-back nature of agent memory.

\paragraph{Retrieval-Augmented Generation.}
RAG~\citep{lewis2020rag} augments an LLM with external retrieval over a fixed corpus.
LongRAG~\citep{jiang2024longrag} studies integration with long-context LLMs,
while M-RAG~\citep{wang2024mrag} uses multiple partitions for fine-grained retrieval.
However, standard RAG typically assumes a static knowledge base,
whereas agentic settings require memory that is continuously updated~\citep{gutierrez2025rag2memory}.

\paragraph{Memory-Augmented Generation.}
MAG systems maintain a time-variant memory that evolves via a feedback loop~\citep{zhong2024memorybank,park2023generative}.
Recent work spans multiple architectural categories~\citep{jiang2026anatomy}:
graph-structured memory (MAGMA~\citep{jiang2026magma}, Zep~\citep{rasmussen2025zep}),
hierarchical tiers (MemGPT~\citep{packer2023memgpt}, MemoryOS~\citep{kang2025memoryos},
EverMemOS~\citep{hu2026evermemos}),
entity-centric stores (A-MEM~\citep{xu2025amem}, Mem0~\citep{chhikara2025mem0}),
and episodic/reflective designs (Nemori~\citep{nan2025nemori}).
All of these operate as \emph{external services} that agents call into.
\byterover{} departs from this paradigm by embedding memory operations
directly in the agent's reasoning loop.

\paragraph{Benchmarks and Evaluation.}
LoCoMo~\citep{maharana2024locomo} evaluates long-term conversational memory
across 35 sessions with temporal and causal reasoning questions.
LongMemEval~\citep{wu2024longmemeval} stress-tests memory retention
at context lengths exceeding 100K tokens.
Recent analysis~\citep{jiang2026anatomy} reveals that many benchmarks
risk context saturation---fitting within modern 128K+ windows---and
that lexical metrics (F1, BLEU) systematically diverge from semantic correctness,
motivating the use of LLM-as-a-Judge evaluation~\citep{zheng2023llmjudge}.

\section{Hyperparameter Configuration}
\label{app:hyperparameters}

\begin{table}[h]
\centering
\small
\begin{tabular}{@{}lll@{}}
\toprule
\textbf{Module} & \textbf{Parameter} & \textbf{Value} \\
\midrule
\multirow{4}{*}{Search Index} & MiniSearch version & v7 \\
 & Fields & title ($5\times$), content ($1\times$), path ($1.5\times$) \\
 & Max retrieval results & 32 \\
 & Max content length & 8{,}000 chars \\
\midrule
\multirow{3}{*}{Search Config} & Fuzzy ratio & 0.2 \\
 & Prefix matching & enabled \\
 & Score normalization & $s / (1 + s)$ \\
\midrule
\multirow{3}{*}{Direct Response} & High confidence threshold & 0.93 \\
 & Minimum score & 0.85 \\
 & Minimum gap (top vs \#2) & 0.08 \\
\midrule
\multirow{2}{*}{OOD Detection} & Minimum relevance score & 0.6 \\
 & Unmatched term threshold & 0.85 \\
\midrule
\multirow{3}{*}{Lifecycle} & Importance decay (daily) & $0.995^{\Delta t}$ \\
 & Recency decay & $e^{-\Delta t / 30}$ \\
 & Maturity thresholds & draft $<$ 35, core $\geq$ 85 \\
\midrule
\multirow{2}{*}{Curation} & Max files per operation & 5 \\
 & Max chars per file & 40,000 \\
\midrule
\multirow{2}{*}{Compression} & Level 1 (LLM) & Normal summarization \\
 & Level 2 (Aggressive) & $0.6\times$ token budget \\
\midrule
\multirow{2}{*}{Cache} & Query cache TTL & 0 (disabled) \\
 & Fingerprint cache TTL & 0 (disabled) \\
\midrule
\multirow{2}{*}{Inference} & LLM Backbone (curate/query) & Gemini 3 Flash \\
 & Temperature & 0.0 \\
\midrule
\multirow{4}{*}{Judge} & Model & Gemini 3 Flash \\
 & Max tokens & 8{,}192 \\
 & Thinking budget & 0 (disabled) \\
 & Prompts & Hindsight (verbatim) \\
\midrule
\multirow{4}{*}{Justifier} & Model & Gemini 3.1 Pro \\
 & Max tokens & 32{,}768 \\
 & Thinking budget & low \\
 & Prompts & Hindsight (adapted for LongMemEval) \\
\bottomrule
\end{tabular}
\caption{Hyperparameter configuration for \byterover{} and evaluation pipeline.}
\label{tab:hyperparameters}
\end{table}

\section{Context Tree Entry Example}
\label{app:entry-example}

Figure~\ref{fig:entry-example} shows a complete knowledge entry
as stored on disk in the \contexttree{}.

\begin{figure}[h]
\begin{tcolorbox}[
  colback=backcolor,
  colframe=codegray!50,
  title={\small\texttt{.brv/context-tree/architecture/module\_boundaries/auth\_billing\_cycle.md}},
  fonttitle=\small\ttfamily,
]
{\small
\begin{verbatim}
---
title: Auth-Billing Circular Dependency
tags: [architecture, circular-dependency, tech-debt]
keywords: [auth, billing, import-cycle, tree-shaking]
related:
  - architecture/module_boundaries/auth_service_deps.md
  - tech_debt/prioritization/q1_2026_assessment.md
importance: 82
maturity: validated
recency: 0.91
accessCount: 7
updateCount: 3
createdAt: 2026-02-03T11:20:00Z
updatedAt: 2026-02-15T09:45:00Z
---

## Relations
@architecture/module_boundaries/auth_service_deps.md
@architecture/module_boundaries/billing_integration.md
@tech_debt/prioritization/q1_2026_assessment.md

## Raw Concept
**Task:** Map circular dependency between auth, billing,
and user-management modules after v1.8 release.
**Changes:** PR #847 introduced auth -> billing import.
**Files:** src/auth/middleware.ts, src/billing/subscriptionCheck.ts
**Timestamp:** 2026-02-03T11:20:00Z
**Author:** architecture-agent

## Narrative
### Structure
The dependency cycle forms a triangle:
auth -> billing -> user-management -> auth.

### Rules
Circular deps with runtime imports are severity: high.
Type-only circular imports are severity: low.
\end{verbatim}
}
\end{tcolorbox}
\caption{A complete knowledge entry in the \contexttree{},
  showing the YAML frontmatter with lifecycle metadata,
  explicit relation annotations, raw concept (provenance),
  and narrative (interpreted structure).}
\label{fig:entry-example}
\end{figure}

\end{document}